\documentclass[11pt]{article}
\usepackage{fullname}
\usepackage{fullpage}
\usepackage{epsfig}
\usepackage{times}

\begin{document}
\title{``I'm sorry Dave, I'm afraid I can't do that'':
Linguistics, Statistics, and Natural Language Processing circa 2001\thanks{To appear in the National Research Council
study on the Fundamentals of Computer Science.  This is an April 2003 version.} }
\author{Lillian Lee, Cornell University}
\date{}
\maketitle

\bibliographystyle{fullname}

\begin{quote}
{\em It's the year 2000, but 
where are the flying cars?  I was promised flying cars.} \\ \hspace*{2.5in} -- Avery
Brooks, IBM commercial
\end{quote}
According to many pop-culture visions of the future, technology
will eventually produce the Machine that Can Speak to Us.  Examples
range from the False Maria in Fritz Lang's 1926 film {\em Metropolis}
to {\em Knight Rider}'s KITT (a {\em talking} car) to {\em Star Wars}'
C-3PO (said to have been modeled on the False Maria).  And, of course,
there is the HAL 9000 computer from {\em 2001: A Space Odyssey}; in
one of the film's most famous scenes, the astronaut Dave asks HAL
to open a pod bay door on the spacecraft, to which HAL responds, ``I'm
sorry Dave, I'm afraid I can't do that''.

Natural language processing, or NLP, is the field of computer science
devoted to creating such machines --- that is, enabling computers to
use human languages both as input and as output.  The area is quite
broad, encompassing problems ranging from simultaneous multi-language
translation to advanced search engine development to the design of
computer interfaces capable of combining speech,
diagrams, and other modalities simultaneously.  A natural consequence of this wide range of
inquiry is the integration of ideas from computer science with work
from many other fields, 
including 
{\em linguistics}, which provides models of language;
{\em psychology}, which provides models of cognitive processes;
{\em information theory}, which provides models of communication; and
{\em mathematics and statistics}, which provide tools for analyzing
and acquiring such models.

The interaction of these ideas together with advances in machine
learning (see [other chapter]) has resulted in concerted research
activity in {\em statistical natural language processing}: making
computers language-enabled by having them acquire linguistic
information directly from samples of language itself.  In this essay,
we describe the history of statistical NLP; the twists
and turns of the story serve to highlight the sometimes complex
interplay between computer science and other fields.

Although currently a major focus of research, the data-driven,
computational approach to language processing was for some time held
in deep disregard because it directly conflicts with
another commonly-held viewpoint:
human language is so complex that language samples alone seemingly
cannot yield enough information to understand it.  Indeed, it is often
said that NLP is ``AI-complete'' (a pun on NP-completeness; see [other
chapter]), meaning that the most difficult problems in artificial
intelligence manifest themselves in human language phenomena. This
belief in language use as the touchstone of intelligent behavior dates
back at least to the 1950 proposal of the Turing Test\footnote{Roughly
speaking, a computer will have passed the Turing Test if it can engage
in conversations indistinguishable from that of a human's.} as a way
to gauge whether machine intelligence has been achieved; as Turing
wrote, ``The question and answer method seems to be suitable for
introducing almost any one of the fields of human endeavour that we
wish to include''.

The reader might be somewhat surprised to hear that language
understanding is so hard.  After all, human children get the hang of
it in a few years, word processing software now corrects (some of) our
grammatical errors, and TV ads show us phones capable of effortless
translation. One might therefore be led to believe that HAL is just
around the corner.

Such is not the case, however. In order to appreciate this point, we
temporarily divert from describing statistical NLP's history --- which touches upon Hamilton versus Madison, the
sleeping habits of colorless green ideas, and what happens when one
fires a linguist --- to examine a few examples illustrating
why understanding human language is such a difficult problem.

\section*{Ambiguity and language analysis}

\begin{quote}
{\em At last, a computer that understands you like your mother.}\\
\hspace*{2.5in} -- 1985 McDonnell-Douglas ad
\end{quote}

The snippet quoted above indicates the early confidence at
least one company had in the feasibility of getting computers to
understand human language.
But in fact, that very sentence is illustrative of the host of
difficulties that arise in trying to analyze human utterances, and so,
ironically, it
is quite unlikely that  the system being promoted
would have been up to the task.  A moment's reflection reveals
that the sentence admits at least three different interpretations:
\begin{enumerate}
  \item The computer understands you as well as your mother understands
  you.
  \item  The computer understands that you like your mother. 
  \item  The computer understands you as well as it understands your mother.
\end{enumerate}
That is, the sentence is {\em ambiguous}; and yet we humans seem to
instantaneously rule out all the alternatives except the first (and
presumably the intended) one.
We do so based on a great deal of background knowledge, including
understanding what advertisements typically try to convince us of.
How are we to get such information into a computer? 

A number of other types of ambiguity are also lurking here.  For
example, consider the speech recognition problem: how can we
distinguish between this utterance, when spoken, and ``... a computer
that understands your lie cured mother''?  We also have a word sense
ambiguity problem: how do we know that here ``mother'' means ``a
female parent'', rather than the Oxford English Dictionary-approved
alternative of ``a cask or vat used in vinegar-making''? Again, it is
our broad knowledge about the world and the context of the remark that
allows us humans to make these decisions easily.

Now, one might be tempted to think that all these ambiguities arise
because our example sentence is highly unusual (although the ad
writers probably did not set out to craft a strange sentence).  Or,
one might argue that these ambiguities are somehow artificial because
the alternative interpretations are so unrealistic that
an NLP system could easily filter them out.  But ambiguities crop up in many
situations.  For example, in ``Copy the local patient files to disk''
(which seems like a perfectly plausible command to issue to a computer),
is it the patients or the files that are local?\footnote{Or, perhaps,
the files themselves are patient?  But our knowledge about the world
rules this possibility out.}  Again, we need to
know the specifics of the situation in order to decide.  And in
multilingual settings, extra ambiguities may arise.  Here is a
sequence of seven Japanese characters:
\begin{center}
\psfig{figure=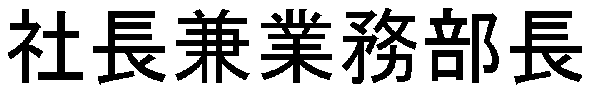,width=1.7in}
\end{center}
Since Japanese doesn't have spaces between words, one is faced with
the initial task of deciding what the component words are. In
particular, this character sequence corresponds to at least two
possible word sequences, ``president, both, business,
general-manager'' (= ``a president as well as a general manager of
business'') and ``president, subsidiary-business, Tsutomu (a name),
general-manager'' (= ?).
It requires a fair bit of linguistic
information to choose the correct alternative.\footnote{To take an
analogous example in English, consider the 
non-word-delimited sequence of letters
``{theyouthevent}''.   This
corresponds to the word sequences ``the youth event'', ``they out he
vent'', and ``the you the vent''.}

To sum up, we see that the NLP task is highly daunting, for to
resolve the many ambiguities that arise in trying to analyze even a
single sentence requires deep knowledge not just about language but
also about the world. And so when HAL says, ``I'm afraid I can't do
that'', NLP researchers are tempted to respond, ``I'm afraid you might
be right''.

\section*{Firth things first}

But before we assume that the only viable approach to NLP is a massive
knowledge engineering project, let us go back to the early approaches
to the problem.  In the 1940s and 1950s, one prominent trend in
linguistics was explicitly empirical and in particular distributional,
as exemplified by the work of Zellig Harris (who started the first
linguistics program in the USA). The idea was that
correlations (co-occurrences) found in language data are important
sources of information, or, as the influential linguist J. R. Firth
declared in 1957, ``You shall know a word by the company it keeps''.

Such notions accord quite happily with ideas put forth by Claude
Shannon in his landmark 1948 paper establishing the field of
information theory; speaking from an engineering perspective, he
identified the probability of a message's being chosen from among
several alternatives, rather than the message's actual content, as its
critical characteristic.  Influenced by this work, Warren Weaver in
1949 proposed treating the problem of translating between languages as
an application of cryptography (see [other chapter]), with one
language viewed as an encrypted form of another.  And, Alan Turing's
work on cracking German codes during World War II led to the
development of the Good-Turing formula, an important tool for
computing certain statistical properties of language.

In yet a third area, 1941 saw the statisticians Frederick Mosteller
and Frederick Williams address the question of whether it was
Alexander Hamilton or James Madison who wrote some of the pseudonymous
Federalist Papers. Unlike previous attempts, which were based on
historical data and arguments, Mosteller and Williams used the
patterns of word occurrences in the texts as evidence. This work led
up to the famed Mosteller and Wallace statistical study which many
consider to have settled the authorship of the disputed papers.

Thus, we see arising independently from a variety of fields the idea
that language can be viewed from a data-driven, empirical perspective
--- and a data-driven perspective leads naturally to a computational
perspective.

\section*{A ``C'' change}

However, data-driven approaches fell out of favor in the late 1950's.
One of the commonly cited factors is a 1957 argument by linguist (and
student of Harris) Noam Chomsky, who believed that language behavior
should be analyzed at a much deeper level than its surface
statistics.  He claimed, 
\begin{quote}
  It is fair to assume that neither sentence (1) [Colorless green
  ideas sleep furiously] nor (2) [Furiously sleep ideas green
  colorless] ... has ever occurred .... Hence, in any [computed]
  statistical model ... these sentences will be ruled out on identical
  grounds as equally ``remote'' from English.\footnote{Interestingly,
  this claim has become so famous as to be self-negating, as simple
  web searches on ``Colorless green ideas sleep furiously'' and its
  reversal will show.} Yet (1), though nonsensical, is grammatical,
  while (2) is not.
\end{quote}
That is, we humans know that sentence (1), which at least obeys (some)
rules of grammar, is indeed more probable than (2), which is just word
salad; but (the claim goes), since both sentences are so rare, they
will have identical statistics --- i.e., a frequency of zero --- in
any sample of English.  Chomsky's criticism is essentially that
data-driven approaches will always suffer from a lack of data, and
hence are doomed to failure.

This observation turned out to be remarkably prescient: even now, when
billions of words of text are available on-line, perfectly reasonable
phrases are not present. Thus, the so-called {\em sparse data problem}
continues to be a serious challenge for statistical NLP even
today. And so, the effect of Chomsky's claim, together with some
negative results for machine learning and a general lack of computing
power at the time, was to cause researchers to turn away from
empirical approaches and toward {\em knowledge-based} approaches where
human experts encoded relevant information in computer-usable form.

This change in perspective led to several new lines of fundamental,
interdisciplinary research.  For example, Chomsky's work viewing
language as a formal, mathematically-describable object has had
lasting impact on both linguistics and computer science; indeed, the
{\em Chomsky hierarchy}, a sequence of increasingly more powerful
classes of grammars, is a staple of the undergraduate computer science
curriculum.  Conversely, the highly influential work of, among others,
Kazimierz Adjukiewicz, Joachim Lambek, David K. Lewis, and Richard Montague
adopted the {\em lambda calculus}, a fundamental concept in the study
of programming languages, to model the semantics of natural languages.

\section*{The empiricists strike back}

By the '80s, the tide had begun to shift once again, in  part
because of the work done by the speech recognition group at IBM.  These
researchers, influenced by ideas from information theory, explored the
power of probabilistic models of language combined with access to much
more sophisticated algorithmic and data resources than had previously been
available. In the realm of speech recognition, their ideas form the
core of the design of modern systems; and given the recent successes
of such software --- large-vocabulary continuous-speech recognition
programs are now available on the market --- it behooves us to examine
how these systems work.

Given some acoustic signal, which we denote by the variable $a$, we
can think of the speech recognition problem as that of transcription:
determining what sentence is most likely to have produced $a$.
Probabilities arise because of the ever-present problem of ambiguity:
as mentioned above, several word sequences, such as ``your lie cured
mother'' versus ``you like your mother'', can give rise to similar
spoken output.  Therefore, modern speech recognition systems
incorporate information both about the acoustic signal and the
language behind the signal.  More specifically, they rephrase the
problem as determining which sentence $s$ maximizes the product
$P(a|s)\times P(s)$. The first term measures how likely the acoustic
signal would be if $s$ were actually the sentence being uttered
(again, we use probabilities because humans don't pronounce words the
same way all the time).  The second term measures the probability of
the sentence $s$ itself; for example, as Chomsky noted, ``colorless
green ideas sleep furiously'' is intuitively more likely to be uttered
than the reversal of the phrase.  It is in computing this second term,
$P(s)$, where statistical NLP techniques come into play, since
accurate estimation of these sentence probabilities requires
developing probabilistic models of language.  These models are
acquired by processing tens of millions of words or more.
This is by no means a simple procedure; even linguistically naive
models require the use of sophisticated computational and statistical
techniques because of the sparse data problem foreseen by Chomsky.
But using probabilistic models, large datasets, and powerful learning
algorithms (both for $P(s)$ and $P(a|s)$) has led to our achieving the
milestone of commercial-grade speech recognition products capable of
handling continuous speech ranging over a large vocabulary.

But let us return to our story. Buoyed by the successes in speech
recognition in the '70s and '80s (substantial performance gains over
knowledge-based systems were posted), researchers began applying
data-driven approaches to many problems in natural language
processing, in a turn-around so extreme that it has been deemed a
``revolution''.  Indeed, now empirical methods are used at all levels
of language analysis.  This is not just due to increased resources: a
succession of breakthroughs in machine learning algorithms  has
allowed us to leverage existing resources much more effectively.
At the same time, evidence from psychology shows that human learning
may be more statistically-based than previously thought; for instance,
work by Jenny Saffran, Richard Aslin, and Elissa Newport reveals that
8-month-old infants can learn to divide continuous speech into word
segments based simply on the statistics of sounds following one
another.  Hence, it seems that the ``revolution'' is here to stay.

Of course, we must not go overboard and mistakenly conclude that the
successes of statistical NLP render linguistics irrelevant (rash
statements to this effect have been made in the past, e.g., the
notorious remark, ``Every time I fire a linguist, my performance goes
up'').  The information and insight that linguists, psychologists, and
others have gathered about language is invaluable in creating
high-performance broad-domain language understanding systems; for
instance, in the speech recognition setting described above, a better
understanding of language structure can lead to better language
models.  Moreover, truly interdisciplinary research has furthered our
understanding of the human language faculty.  One important example of
this is the development of the {\em head-driven phrase structure
grammar} (HPSG) formalism --- this is a way of analyzing natural
language utterances that truly marries deep linguistic information
with computer science mechanisms, such as unification and recursive
data-types, for representing and propagating this information
throughout the utterance's structure.  In sum, computational
techniques and data-driven methods are now an integral part both of
building systems capable of handling language in a domain-independent,
flexible, and graceful way, and of improving our understanding of
language itself.

\subsection*{Acknowledgments} Thanks to the members of the CSTB
Fundamentals of Computer Science study --- and especially Alan
Biermann --- for their helpful feedback.  Also, thanks to Alex Acero,
Takako Aikawa, Mike Bailey, Regina Barzilay, Eric Brill, Chris
Brockett, Claire Cardie, Joshua Goodman, Ed Hovy, Rebecca Hwa, John
Lafferty, Bob Moore, Greg Morrisett, Fernando Pereira, Hisami Suzuki,
and many others for stimulating discussions and very useful comments.
Rie Kubota Ando provided the Japanese example.  The use of the term
``revolution'' to describe the re-ascendance of statistical methods
comes from Julia Hirschberg's 1998 invited address to the American
Association for Artificial Intelligence.  I learned of the
McDonnell-Douglas ad and some of its analyses from a class run by
Stuart Shieber.  All errors are mine alone.  This paper is based upon
work supported in part by the National Science Foundation under ITR/IM
grant IIS-0081334 and a Sloan Research Fellowship.  Any opinions,
findings, and conclusions or recommendations expressed above are those
of the authors and do not necessarily reflect the views of the
National Science Foundation or the Sloan Foundation.

\section*{For further reading}
\newcommand{\myind}{\hspace*{.3in}}

\noindent Charniak, Eugene.
\newblock 1993.
\newblock {\em Statistical Language Learning}.
\newblock MIT Press.

\bigskip

\noindent Jurafsky, Daniel and James~H. Martin.
\newblock 2000.
\newblock {\em Speech and Language Processing: An Introduction to Natural
 \\ \myind  Language Processing, Computational Linguistics, and Speech Recognition}.
\newblock Prentice Hall.
\newblock Contribut-  \\ \myind ing writers: Andrew Kehler, Keith Vander Linden, and Nigel
  Ward.

\bigskip

\noindent Manning, Christopher~D. and Hinrich Sch\"{u}tze.
\newblock 1999.
\newblock {\em Foundations of Statistical Natural Language 
Process-\\ \myind ing}. The MIT Press.

\end{document}